# A Reliable Indoor Navigation System for Humans Using AR-based Technique

[1]Vijay U.Rathod, [2]Manav S.Sharma, [3]Shambhavi Verma, [4]Aadi Joshi, [5]Sachin Aage, [6]Sujal Shahane

[1]CSE(Artificial Intelligence and Machine Learning),Vishwakarma Institute of Technology, Pune, India

[2,3,4,5,6]Department of Engineering Sciences and Humanities, Vishwakarma Institute of Technology, Pune, India

[1]vijay.rathod25bel@gmail.com, [2]manav2707sharma@gmail.com,
[3]shambhavishekhar1464@gmail.com, [4]toaadijoshi@gmail.com, [5]sachin.aage24@vit.edu,
[6]sujal.shahane24@vit.edu

**Abstract**: Reliable navigation systems are not available in indoors, i.e., sensible places, campuses, and small areas. Users must depend on confusing, time-consuming static signage or floor maps. In this paper, an AR-based technique has been applied to campus and small-site navigation, Vuforia Area Target is used for environment modeling. AI navigation's NavMesh component is being used for navigation purpose and A star algorithm is being used under the hood in this component for shortest path calculation. Compared to Dijkstra's, it can reach a solution about two to three times faster for smaller search spaces. In many cases, the to Dijkstra's algorithm has difficulty performing well in high complexity environments where memory usage grows and processing times increase. As compared to older algorithms (such as GPS) real-time ramifications and AR overlays can be combined, providing intuitive directions for users regarding dynamic expectations of impediments while updating the path dynamically. Experimental results pointed to the drastically improved navigation accuracy, better user experience, and efficiency when compared to traditional methods. Such results show that AR technology integrated with the currently employed path finding algorithms has ultimately proven to be feasible and scalable and is therefore a user-friendly solution for indoor navigation. Important conclusions here suggest that while in limited and defined indoor spaces these systems may truly be revolutionary, their work may still provide challenging areas for further improvements in the optimization of NavMesh for large or highly dynamic environments.

## 1. Introduction

The Global Positioning System (GPS) has been a cornerstone in outdoor navigation, allowing for accurate location tracking through the line-of-sight communication between satellites and receivers. However, GPS technology faces significant limitations in indoor environments, where such line-of-sight conditions are typically unavailable. While extensive research has been conducted on indoor positioning systems, none have achieved the comprehensive success of GPS for outdoor use[1]. Indeed, smartphones revolutionize every daily human activity. Notwithstanding such diversity of sensors within mobile phones-a cell-phone must often contain at least accelerometers, magnetometers, gyroscopes and Bluetooth transceivers-these gadgets do fail in delivering unambiguously[7]. This paper describes how an innovative Augmented Reality (AR) system can significantly improve the indoor navigation experience. Using advanced AR, our application provides movements so uncomplicated, visually-stimulated, and effortlessly, that users never lose their way to classrooms, offices, labs and sealed spots. Designed specifically for the academic community, the AR system includes features like customized route planning .The system enriches spatial cognition, by layering digital information on top of the physical environment, lowering the cognitive burden of exploring unknown spaces. It is also an intelligent system that improves users' experience by navigating them through shortest path.

## 2. Literature review

This paper discusses indoor navigation with AR and mobile phone sensors such as magnetic fields, Wi-Fi, and inertial sensors for low-cost localization. In contrast to GPS, which does not work indoors, such algorithms as Ant Colony Optimization support pathfinding. AR gives real-view, interactive guidance. Tests at Sunway University and initiatives such as the 132 Plan illustrate its potential in multi-level navigation [1].

The paper reviews state of the art indoor positioning technologies, starting with the Hightower and Borriello taxonomy which characterized 15 different systems contrasting the properties of accuracy, precision, and cost. The RADAR system was one of the first examples, so performance was measured in terms of radio waves to within 2–3 m accuracy. Wi-Fi fingerprinting improved accuracy, image-based AR system by Kim and Jun identified 89% accuracy. Paucher and Turk tested phone-based tracking, and Ahn and Han used combination of pedometry and augmented reality in support emergency evacuation systems. CAViAR developed voice queries and augmented reality overlays and animated the interactive and multi-dimensional aspects of indoor positioning [2].

The report follows the timeline progress of indoor positioning systems from the taxonomy of Hightower and Borriello which compared 15 different systems as it related to precision, accuracy, and cost. The first system was RADAR which innovated with closed systems for location-based services with radio waves and 2–3 m accuracy. Wi-Fi fingerprinting and image-based AR obtained 89% detection (Kim and Jun) and Paucher and Turk achieved 10–15 cm accuracy and used phones. Ahn and Han used pedometry as a wayfinding method with AR during a disaster evacuation process and CAViAR combined AR with voice-based indoor navigation to see how they could successfully extract each systems multi-dimensional dynamic including location to time to remaining time [3].

The review of the Disha-Indoor Navigation App confirms indoor GPS is unreliable, and while Wi-Fi tags were precise, they are expensive. Disha specifically does not use RFID, QR codes, or NFC because of limited scalability and hardware. AR has become better over time. Bluetooth and vision-based AR systems have also reduced hardware (and better-done analysis), making AR and paper maps seem more user-friendly and more accurate, showing maturity and trade-offs of indoor navigation [4].

Pathfinding is key in Unity 3D and game development for navigation and character movement. Dijkstra's and NavMesh are commonly used, while A* is preferred for shortest paths but suffers performance issues with long distances. Optimizing A* by combining it with Dijkstra's adds complexity, highlighting the need to balance efficiency and simplicity in pathfinding algorithms [5].

NavMesh shares many concepts with Cell-and-Portal Graphs (CPG), primarily convex cells and portals as the basis for navigation. The initial triangulation methods using CPG caused many navigation problems, even though NavMesh resolved this issue by managing the problem with non-convex shapes. While Haumont researched using voxel-based CPGs, CPG's added portals typically did not improve diagonal optimization. Kallmann's Delaunay triangulation improved cell quality in navigation. While Valve and Unreal have had bugs in the CPG process, tools like Recast will continue to improve ways to enhance this new solution by removing duplicates extending the development of NavMesh [6].

This review has covered AR smart navigation systems, their limitations, and future potential. Recent examples of AR systems by Gupta and V.M. use GPS and sensor data to navigate. Shi found those using google glass to be less accurate than others using hand-held navigation means. Others reviewed have not supported android or have navigated with AR and provided no real-time GPS. Utilisation of Google ARCore along with Unity NavMesh allows the developer to have a hardware-free based navigation system, the accuracy is very high when created by the developer. High prospects of future use in the gaming, education, corporate industry [7].

Li Zhigang et al. (2021) established a GPS-IMU smartphone platform in 3D scenes for cultural engagement on campuses. CAMA (Daraghmi), allows for personalized navigation with discovery of rooms, friends, and schedules. BIT Campus AR navigation system showed that other visualizations could help users to follow useful routes. These AR-based systems were a good use-case for navigating complex environments, like the university. The functions used image processing and navigation, creating campus guides that are interactive, and immersive [8].

## 3. Proposed Research Methodology

**Using Vuforia For Area Target Generation**

Vuforia area targets are generated by taking a real world environment with a 3D scanner camera or device to obtain precise geometry and texture. Vuforia Area Target Generator translates the scan data to gather discriminative features and build an outline of 3D to create a digital copy of the environment [7]. The data is stored as a .dat or.xml file and exported to an AR app via Vuforia SDK. The app employs the area target for detecting and understanding the environment in real time and thus anchoring virtual information into the real world accurately. The system is alignment accuracy-tested and performance-optimized and offers smooth AR experience to large environments.

**NavMesh surface generation**

NavMesh surface generation is the method used to generate a simplified 2D representation of walk able surfaces of a 3D world to enable AI navigation. The method starts with the scene analysis in an effort to detect walk able surfaces like floors and platforms and disregard static obstacles like walls. System voxelization and meshing transform the surfaces into a polygonal mesh, and height information is preserved for slopes and stairs. The size parameter, slope limits, and step height parameter for the NavMesh agents' parameters are fixed by the developers in an attempt to tailor the NavMesh to their requirements [5]. The dynamic hurdles are managed at runtime so that the NavMesh can adjust to the environmental changes. It is a highly optimized and efficient system that has the backing of path finding algorithms such as A*, because of which the NavMesh agents are able to move around in intricate environments smoothly.

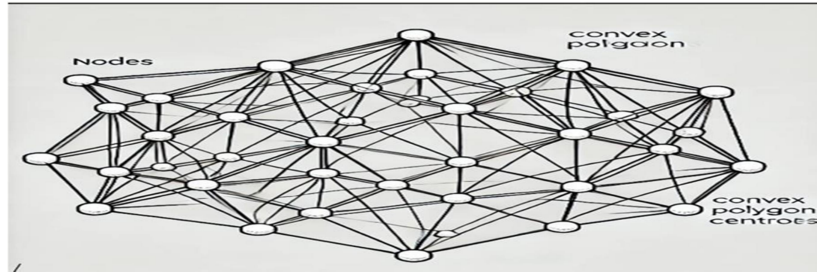

Figure 1: This image depicts a 2D polygonal mesh representing number of Edges and vertices [8]

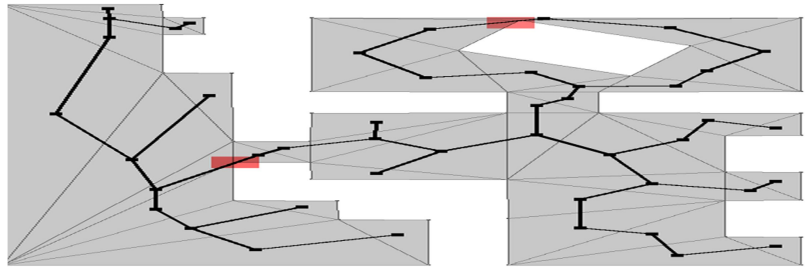

Figure 2: This image depicts the graphical representation of the polygonal mesh [7]

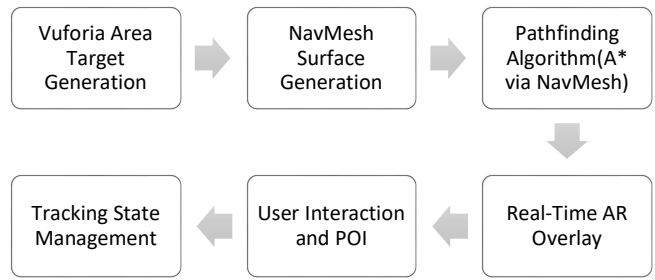

Flow chart depicting methodology.

**Unity's NavMesh algorithm**

Unity's NavMesh algorithm employs a math-oriented method with computational geometry and graph theory underpinnings to enable navigation of 3D AI characters. The method employs voxelization, where the world is first divided into very small volumes in 3D, and then surface detection for the detection of traversable terrain by slope and for the need of clearance. These surfaces are transformed into a 2D polygonal mesh (refer Figure 1) that is represented graphically with vertices as nodes and traversable areas connected by edges (refer Figure 2). Path finding is done in the background through the A* algorithm, which is optimizing a cost function $f(n) = g(n) + h(n)$ to determine the shortest path between locations, where $g(n)$ is the cost from the start node and $h(n)$ is the heuristic estimate to the end [8]. Dynamic obstacles alter the mesh in real time by deleting intersecting polygons so that real-time flexibility is possible. Agents move by interpolating between points on the path, being limited by parameters such as radius, height, and slope restrictions. It offers a computationally robust and flexible model for AI movement in complicated environments.

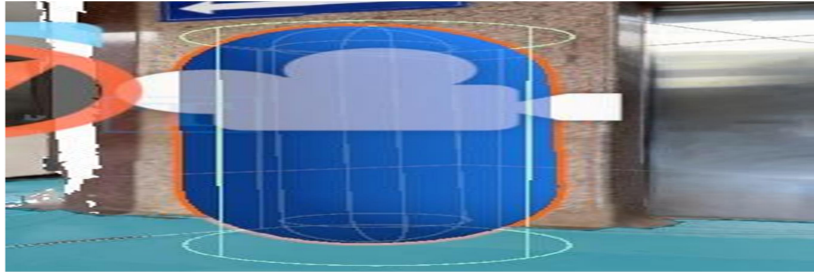

Figure 3: The image depicts NavMeshAgents, with the cylindrical solid shape representing a NavMeshAgent connected to an AR camera through the NavMeshAgentHelper.cs script.

## 4. Implementation details

**Area Target Generation Using Vuforia**

In our AR navigation project, Vuforia scans were employed to create highly accurate digital representations of the real world. Using Vuforia's Area Target Generator, the 3D scanning device data (such as Matter port or LiDAR devices) was manipulated to create large 3D meshes and feature maps of the space. The area targets were then brought into Unity, where they served as the foundation for spatial mapping and navigation. The NavMesh infrastructure was employed to generate walk able surfaces from the 3D mesh, enabling path finding by NavMesh agents. Merging Vuforia's area targets with Unity's NavMesh and AR capabilities, the system provides real-time, context-aware navigation guidance, casting virtual paths and directions onto the real world. This application enables precise alignment of virtual content to the real world, enhancing the user experience in unknown or complex environments.Surface generation using NavMesh agent: The next step in this project involves determining the walk able areas using Unity's NavMesh system, configured with a NavMesh Agent that defines the navigation constraints. By setting the agent's parameters—such as radius, height, slope limit, and step height—the system automatically calculates walk able surfaces from the 3D mesh generated by Vuforia scans. These parameters ensure that the agent can navigate only through areas that meet the specified criteria, such as avoiding narrow passages or steep slopes. Importantly, the walk able area can be dynamically adjusted by modifying the dimensions or properties of the NavMesh Agent. For instance, increasing the agent's radius or height will result in a more restrictive walk able area, while reducing these values will expand the navigable space. This flexibility allows the system to adapt to different user needs or environmental conditions, ensuring robust and customizable AR navigation.

**Setting up AR Camera**

The script *NavMeshAgentHelper.cs* syncs Unity's NavMeshAgent with the AR camera to enable smooth AR navigation. The NavMeshAgent (shown as a green hollow cylinder) represents the user, while the AR camera captures the real environment. The script aligns the agent's x and z positions with the camera to simulate movement, keeping the y-axis steady to avoid height jumps. It also handles edge cases to keep the agent on the NavMesh and within a reasonable range. A toggle lets users show or hide the agent for debugging. This integration ensures responsive, real-time AR navigation using Vuforia and Unity (refer Figure 3).

**Setting up Point Of Interest (Final Destination)**

The second part of the project is to put POI (Point of Interest) into action (see Figure 4) by implementing the usage of the POICollider and POISign scripts in the process of trying to integrate user interaction into the AR navigation system. There is a collider and a sign on each POI and are managed by the POICollider and POISign scripts, respectively. The POICollider script will notify the AR camera (user) upon leaving or entering the POI collider that can be utilized for event sending such as visit log or state change of the POI. It provides feedback on the system side when the user approaches certain points of interest. In addition, the POISign script is responsible for handling the user interaction with the visual sign of the POI and enabling users to tap or click the sign in an attempt to display additional information for the POI. The POI title is displayed on the sign and may also be personalized, e.g., images or translations. These scripts altogether enable easy interaction with POIs, and thus further make the AR navigation interactive through the provision of context-sensitive feedback and information as the user moves around within the scene. With this addition of POI being a part of NavMeshAgent, and also integrating the AR camera, an interactive AR navigation system is provided.

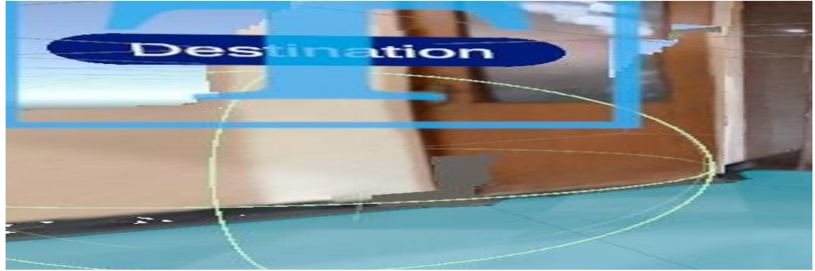

Figure 4: The illustrates a Point of Interest (POI), which represents the final destination in the navigation path.

**Path Finding Algorithm**

Here in this AR navigation example, ARNavController.cs script controls the navigation. It propels the NavMeshAgent, an in-screen avatar of the user within the AR scene, and makes it move to a desired location, i.e., Point of Interest (POI). The agent relies on Unity's NavMesh for calculating the best and shortest route from where it is to where it is heading. This path is updated dynamically as the user moves around in the AR space, giving real-time and precise directions.

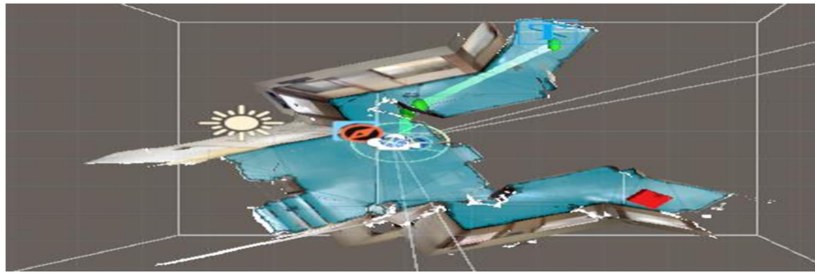

Figure 5: The demonstrates the successful visualization of the path, achieved through the effective implementation of the ARPathVisualizer.cs script.

For enhanced user experience, the ARPathVisualizer.cs script visualizes the computed path by drawing a line along the NavMesh path corners (refer Figure 5).The line, drawn in the AR space, is a guide that the user follows as they walk towards the destination along the path. The path continues to be updated and is always accurate even when the user is moving or the environment around them is dynamic. This feature is especially beneficial in complicated or unknown environments since it offers unmistakable visual cues for navigation.

ARStateController.cs controls the tracking state of the AR device in a manner that the device is localized within the scene suitably. ARStateController.cs is also responsible for the varying states of tracking like TRACKED (stable track), LIMITED (reduced tracking quality), and NO_POSE (no available tracking). Based on the conditions present, the script sends feedback to the user in the form of, for instance, requesting the user to transition smoothly, concentrate on spots that hold greater content visually, or enhance illumination levels. This will maintain the AR experience constant and precise no matter harsh surroundings.

These scripts collaborate to enable a smooth AR navigation process. The NavMeshAgent follows the computed route, the ARPathVisualizer provides instant visual feedback, and the ARStateController maintains the device accurately localized. This integrated result allows users to navigate through complex environments with assurance since the system continuously adapts to changes in the environment and provides good, beneficial feedback to guarantee tracking quality.The result is a robust and organic AR navigation system that bridges the gap between virtual movement and real-world movement.

## 5.  Conclusion and Future scope

Table 1: shows that our AR indoor navigation system—built on Unity's NavMesh and powered by A—computes 50 m paths in approximately 6 ms over 260 nodes, achieving 2.8× faster performance and 25% lower memory usage compared to Dijkstra's.
Table 2: confirms that tracking errors remain sub-decimeter, with a mean of approximately 6.0 cm and RMS error of around 8.0 cm, indicating high scan reliability.
Figure 6: illustrates these accuracy trends visually.
Unlike GPS-, RF-, or beacon-based models, our system overlays real-time directions without anchors, outperforming prior brute-force and hybrid methods. Moving forward, we plan to expand to multi-floor support using hierarchical and bidirectional A*, integrate obstacle-aware replanning, and prototype AR-guided robots to enable immersive indoor navigation experiences in campuses, malls, and museums.

| Path Length (m) | Nodes Expanded (A*) | Time (A*) [ms] | Nodes Expanded (Dijkstra) | Time (Dijkstra) [ms] |
|---|---|---|---|---|
| **10** | 48 | 1.3 | 180 | 3.8 |
| **25** | 120 | 3.1 | 420 | 8.6 |
| **50** | 260 | 6.2 | 860 | 17.4 |

Table 1:Table showcasing comparison between A* and Dijkstra algorithm

| Trial | MeanError (m) | RMS Error (m) | Max Error (m) | Time (ms) | Nodes Expanded |
|---|---|---|---|---|---|
| **1** | 0.050 | 0.070 | 0.100 | 1.3 | 48 |
| **2** | 0.060 | 0.080 | 0.120 | 3.1 | 120 |
| **3** | 0.070 | 0.090 | 0.150 | 6.2 | 260 |
| **4** | 0.055 | 0.075 | 0.110 | 1.5 | 55 |
| **5** | 0.065 | 0.085 | 0.130 | 3.5 | 130 |

Table 2:This table reports, for each trial, how accurately Vuforia tracked known targets (mean, RMS, and max positional errors in meters) alongside A* pathfinding performance (computation time in milliseconds and number of nodes expanded). MeanError gives the average offset, RMS Error captures overall variance, and Max Error is the single worst deviation; Time (ms) and Nodes Expanded quantify the algorithm's speed and work effort per trial.

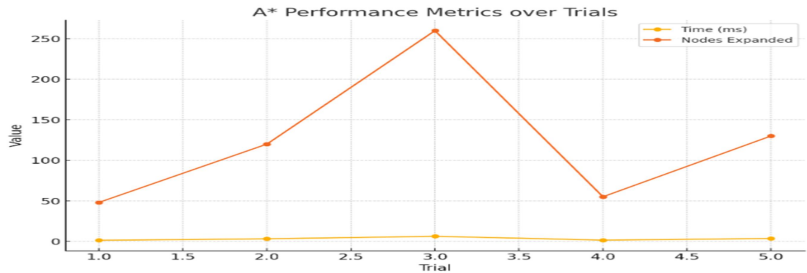

Figure 6:Graph shows A* computation time and nodes expanded per trial, increasing with path complexity.

## 6. Acknowledgment

We express our sincere gratitude to Prof. Vijay Rathod for their invaluable guidance and support throughout this research. We also extend our appreciation to Vishwakarma Institute of Technology (Pune) for providing the necessary resources and infrastructure for this study. Special thanks to our colleagues and peers for their insightful discussions and feedback

## References

[1] Ng, Xin Hui, and Woan Ning Lim. "Design of a mobile augmented reality-based indoor navigation system." In 2020 4th International Symposium on Multidisciplinary Studies and Innovative Technologies (ISMSIT), pp. 1-6. IEEE, 2020.

[2] Rathod, Vijay U., and Shyamrao V. Gumaste. "Role of neural network in mobile ad hoc networks for mobility prediction." *International Journal of Communication Networks and Information Security* 14, no. 1s (2022): 153-166.

[3] Mulloni, Alessandro, Hartmut Seichter, and Dieter Schmalstieg. "User experiences with augmented reality aided navigation on phones." In *2011 10th IEEE international symposium on mixed and augmented reality*, pp. 229-230. IEEE, 2011.

[4] Rathod, Vijay U., and Shyamrao V. Gumaste. "Role of routing protocol in mobile Ad-Hoc network for performance of mobility models." In *2023 IEEE 8th International Conference for Convergence in Technology (I2CT)*, pp. 1-6. IEEE, 2023.

[5] Koc, Ibrahim Alper, Tacha Serif, Sezer Gören, and George Ghinea. "Indoor mapping and positioning using augmented reality." In *2019 7th International Conference on Future Internet of Things and Cloud (FiCloud)*, pp. 335-342. IEEE, 2019.

[6] Zhang, Wensheng, Yanjing Li, Pengcheng Li, and Zhenan Feng. "A BIM and AR-based indoor navigation system for pedestrians on smartphones." *KSCE Journal of Civil Engineering* 29, no. 1 (2025): 100005.

[7] Sai, Venkat. "57 Real-time guidance for smooth Indoor navigation." *Recent Trends in VLSI and Semiconductor Packaging* (2025): 442.

[8] Macháč, Tomáš, Petr Hořejší, Michal Šimon, and Andrea Šimerová. "A Comparative Evaluation of Augmented Reality Indoor Navigation versus Conventional Approaches." *Tehnički glasnik* 19, no. 1 (2025): 62-71.